\documentclass[11pt]{article}
\usepackage{amssymb}
\usepackage{amsmath}
\usepackage{multirow}
\usepackage[margin=2.5cm]{geometry}

\usepackage{graphicx} 
\title{Accurate Water Level Monitoring in AWD Rice Cultivation Using Convolutional Neural Networks
}

\author{Ahmed Rafi Hasan$^{1,3}$, Niloy Kumar Kundu$^{1,3}$,  Saad Hasan$^{2,3}$,\\ Mohammad Rashedul Hoque$^{3,4}$, Swakkhar Shatabda$^{3,5}$
\thanks{swakkhar.shatabda@bracu.ac.bd}\\
$^1$Department of Computer Science and Engineering, United International University\\
$^2$School of Business and Economics, United International University\\
$^3$Nodes Digital Limited\\ 
$^4$Faculty of Business Administration, American International University of Bangladesh
\\$^5$
Department of Computer Science and Engineering, BRAC University
}
\date{}
\begin{document}
\maketitle

\begin{abstract}
The Alternate Wetting and Drying (AWD) method is a rice-growing water management technique promoted as a sustainable alternative to Continuous Flooding (CF). Climate change has placed the agricultural sector in a challenging position, particularly as global water resources become increasingly scarce, affecting rice production on irrigated lowlands. Rice, a staple food for over half of the world's population, demands significantly more water than other major crops. In Bangladesh, \textit{Boro} rice, in particular, requires considerable water inputs during its cultivation. Traditionally, farmers manually measure water levels, a process that is both time-consuming and prone to errors. While ultrasonic sensors offer improvements in water height measurement, they still face limitations, such as susceptibility to weather conditions and environmental factors. To address these issues, we propose a novel approach that automates water height measurement using computer vision, specifically through a convolutional neural network (CNN). Our attention-based architecture achieved an $R^2$ score of 0.9885 and a Mean Squared Error (MSE) of 0.2766, providing a more accurate and efficient solution for managing AWD systems.\end{abstract}


% \begin{highlights}
% \end{highlights}



\section{Introduction}
\label{sec:sample1}
Irrigation is the largest water-demanding sector in Bangladesh, accounting for approximately 86\% of the total water use, primarily for crop production \cite{mojid2021water}. This demand is closely tied to regional climatic conditions. As one of the most vulnerable countries to climate change, Bangladesh faces critical challenges from water-related impacts \cite{agrawala2003development}. Observations and climate projections indicate that freshwater resources will be severely affected by climate change, with far-reaching consequences for human health and agriculture \cite{jimenez2014freshwater}. According to the IPCC, by 2050, about 25\% of the global population will live in regions facing water scarcity due to climate change and increasing water demand \cite{thefinancialexpressClimateChange}. In Bangladesh’s southwest coastal zone, the total freshwater river area is projected to shrink from 40.8\% in 2012 to 17.1\% by 2050 \cite{tbsnewsWhatDoes}.

Bangladesh’s crop production spans three concurrent seasons, with rice (paddy) covering about 42\% of the land during the Boro season. This season accounts for 19.56 million tons of clean rice, around 54\% of the country’s total rice production \cite{googleAgricultureYearboook}. The Bangladesh Agricultural Yearbook 2021 reports that 97.36\% of Boro rice was cultivated using irrigation, with 93.67\% of that irrigation sourced from groundwater. Groundwater irrigation, particularly for dry-season rice, has expanded significantly in recent decades, playing a crucial role in the country’s food security \cite{mainuddin2015national}. However, this increased consumption has adversely impacted the water balance, especially in the northwest, where groundwater levels have dropped due to excessive usage \cite{kirby2015impact}.

As freshwater availability declines across many Asian countries including Bangladesh, the demand for rice continues to rise \cite{pingali1997asian}. Approximately 50\% of available freshwater is used for rice production \cite{guerra1998producing}, but this resource is no longer as abundant as it once was \cite{bindraban2001water}. The growing demand for rice, coupled with increasing water scarcity, necessitates producing more rice with less water \cite{guerra1998producing}. In the face of climate uncertainties and hydrological extremes, adopting efficient water management practices is critical to sustaining food production for about half the global population. Lampayan et al. \cite{bouman2007water} demonstrated that the water level in experimental fields was manually monitored using a measuring stick on a PVC water tube. Reducing water use in irrigation systems, especially for rice and other economically important crops, is essential.

One promising water-saving technique is the Alternate Wetting and Drying (AWD) irrigation method, where the field is allowed to dry until a threshold water level is reached before re-irrigation \cite{bouman2007water}. AWD has been proven to reduce irrigation water consumption without significantly affecting rice yields \cite{lampayan2015effects,li2004increasing,carrijo2017rice,howell2015alternate}. It also reduces methane (CH4) emissions \cite{wassmann2010rice,linquist2015reducing,lagomarsino2016alternate,chiaradia2015integrated,setyanto2018alternate,balaine2019greenhouse,sander2020potential,hossain2022farmers}, making it a viable option for mitigating global warming potential from rice cultivation. Linquist et al. \cite{linquist2015reducing} found that AWD reduced global warming potential by 45–90\% compared to continuous flooding. Besides environmental benefits, AWD also lowers labor requirements by reducing the frequency of irrigation \cite{rejesus2011impact}. Demonstrations and training programs are key to encouraging its adoption among farmers \cite{li2004increasing,rahman2017application}. Farmers who implemented AWD reduced irrigation frequency by 28\% on average and saw a nearly 20\% reduction in irrigation costs \cite{kurschner2010water}. In some cases, rice yields increased by 0.4 to 0.5 tons per hectare (around 10\%), with farmers reporting healthier crops and improved plant development \cite{kurschner2010water,maniruzzaman2014water,rahman2017application}.

However, excessive standing water can reduce transpiration rates \cite{zhang2008yield}, and decreased stomatal conductance can lead to reduced photosynthesis and biomass \cite{wong1985leaf}. This presents a challenge in manually monitoring field water levels to ensure they do not fall below the safe threshold. Advances in sensor technology have alleviated this burden by enabling remote monitoring. Recent innovations in real-time field water level monitoring have helped farmers make informed irrigation decisions. Wireless sensors, capable of monitoring field water levels, soil moisture \cite{chiu2020development}, and climatic variables \cite{cruz2022low,kawakami2016rice}, have evolved significantly. Kalyan et al. \cite{kalyan2022alternate} suggested using soil moisture sensors to measure water height in AWD, while Pham et al. \cite{pham2021using} demonstrated that IoT-based water level measurements could maximize AWD’s water-saving benefits. These technologies offer various sensor types, communication protocols, and system architectures, providing affordable and effective solutions for real-time water monitoring \cite{cruz2022low}. 

Despite these advancements, ultrasonic sensors used in irrigation systems have limitations, especially during harsh weather, which can reduce accuracy and reliability. Therefore, a more reliable and real-time water height estimation system is needed to ensure accurate measurements over large irrigation areas. 

The primary contributions of this study are as follows:

\begin{enumerate}
\item We introduce a novel dataset specifically designed for water height estimation in AWD systems. This dataset plays a crucial role in improving the accuracy and efficiency of water level monitoring in lowland crop production systems.
   
\item Our study proposes an innovative solution that integrates camera technology with AI algorithms for real-time monitoring of water levels. This approach provides a more accessible and cost-effective alternative to traditional AWD methods, eliminating the need for manual measurement and addressing the limitations of ultrasonic sensors.
   
\item We utilize the ConvNext architecture, enhanced with an attention mechanism, to improve the performance of our water height estimation model. This combination allows the system to focus on relevant features in the input data, leading to more precise water level predictions.
   
\item The proposed solution provides a scalable and reliable alternative to existing AWD technologies. By offering a more efficient and accessible system, it supports sustainable agriculture practices, addressing water scarcity and distribution challenges exacerbated by climate change.

\end{enumerate}

\section{Proposed Methodology}
In this section, we outline the architecture of the proposed network for identifying water height in the AWD system and provide details of its component modules. We begin by briefly describing the architecture of the pre-trained networks we used, ConvNeXt, followed by a detailed explanation of our proposed architecture and the modules integrated within it. Additionally, we incorporate an attention mechanism to enhance the model's focus on relevant features. Figure 1 illustrates the overall workflow of our proposed work.

\begin{figure*} [ht]
  \includegraphics[width=\textwidth]{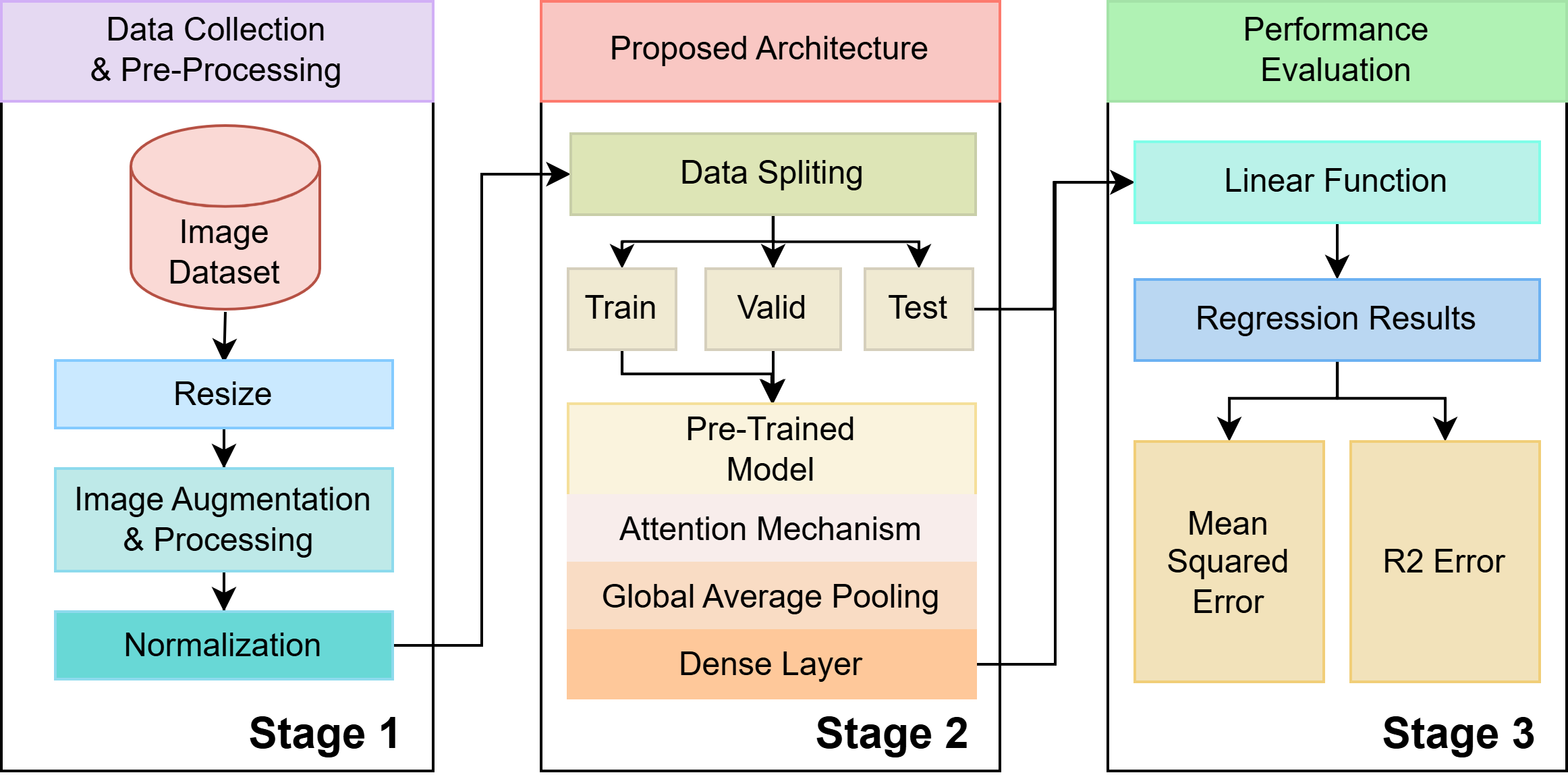}
  \caption{Workflow of our proposed work.}
  \label{fig:workflow}
\end{figure*}

\subsection{Overview of Our Proposed Architechture}

From Figure \ref{fig:workflow}, we start by preprocessing the images. We resize them to  $224 \times 224$ dimensions and normalize their pixel values to ensure they are consistent. After preprocessing, we apply image augmentation techniques, which increase the total number of images from 1,275 to 2,250. Finally, we split the dataset into training, testing, and validation sets in a ratio of 80:10:10.

\begin{figure*}[ht]
\centering
  \includegraphics[width=0.65\textwidth]{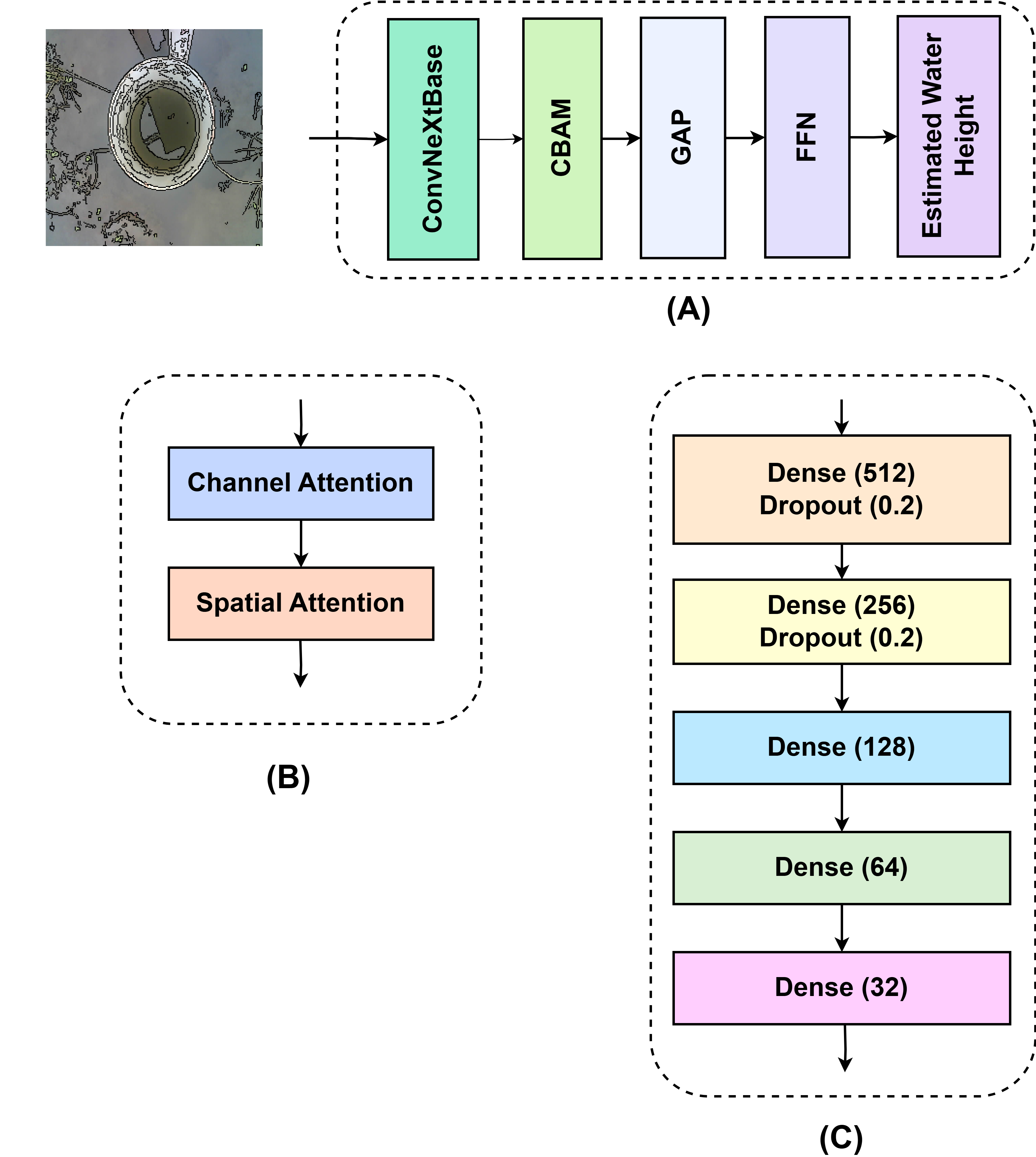}
  \caption{Comprehensive view of the (A) proposed architecture, (B) Convolutional Block Attention Module (CBAM) \cite{woo2018cbam}, and (C) Feed-Forward Network (FFN).}
  \label{fig:model_arch}
\end{figure*}

In the proposed architecture, as depicted in figure \ref{fig:model_arch}(A), the training and validation data are fed into our pre-trained ConvNeXtBase model to utilize its robust feature extraction capabilities. ConvNeXtBase is a state-of-the-art convolutional neural network that incorporates modern design choices, making it highly effective for various image recognition tasks. By using pre-trained weights, we benefit from the knowledge the model has already gained from being trained on large datasets, which can significantly improve our model's performance even with a relatively small dataset.

To further enhance the model's performance, we integrate the Convolutional Block Attention Module (CBAM) \cite{woo2018cbam} attention mechanism, as depicted in figure \ref{fig:model_arch}(B). This mechanism refines the feature maps by focusing on the most relevant parts of the image. This improves the model's ability to distinguish important features from less important ones, aiding in the accurate prediction of water height in the AWD system.

Following this, we apply Global Average Pooling (GAP) to reduce the spatial dimensions of the feature maps, making them suitable for the fully connected layers. GAP reduces each feature map to a single value, helping to reduce the overall number of parameters and prevent overfitting. The output from the GAP layer is then passed through five dense layers with sequentially decreasing neurons: 512, 256, 128, 64, and 32. Each dense layer uses the ReLU activation function to introduce non-linearity and enable the model to learn complex patterns from the data. This sequential reduction in neurons helps in progressively reducing the dimensionality of the data, leading to effective feature extraction and learning.

Finally, in the performance evaluation stage, the processed data is passed through a dense layer with a linear activation function to perform the regression task, as depicted in figure \ref{fig:model_arch}(C). The linear activation function is chosen because regression tasks require predicting continuous values. The model outputs the predicted water height in the AWD system, and we evaluate the regression results using metrics such as MSE, $r^2$ error, and training vs. validation loss, providing critical metrics for optimizing water usage in agricultural practices.

% \subsection{Attention-based model}
% In this section we propose a attention-based model for water height detection in AWD system for large irrigation areas. We experimented different Convolutional Neural Network(CNN) architecture along with the different attention mechanisms to improve the performance of the model. The  proposed 2D attention-based model aims to capture the important features and spatial relationships in the input images, allowing for accurate detection of water height in the AWD system.

\subsection{ ConvNeXt as a Base Architecture}
% % These days, deep learning is a highly popular approach that is applied in many areas of daily life, such as transportation, aviation, agriculture, and illness prediction. There are several uses for pre-trained and deep learning algorithms. In this work, we have experimented different  pre-trained deep learning-based networks called Resnet50, Densenet, InceptionNet, ConvNext etc for transfer learning with the goal of identifying the water height in the AWD system. 

% Transfer learning is a deep learning technique where a model trained on one large dataset is used for a different but related task. This approach uses the knowledge gained from the initial training, allowing the model to perform well even with limited data for the new task. Using pre-trained models saves time and computational resources while also taking advantage of their strong feature extraction abilities.

ConvNeXt \cite{liu2022convnet} is a state-of-the-art architecture built upon the original ResNet50 framework and incorporates ideas from hierarchical Vision Transformers, like the Swin Transformer \cite{liu2021swin}. It has a multistage architecture with different resolutions for feature maps at each stage. Key design aspects of ConvNeXt include the stage compute ratio (SCR) and the structure of the stem cell. The SCR, with a ratio of 3:3:9:3, defines the number of blocks per stage. The stem cell structure is updated with a patchify layer that uses a 4x4 convolutional kernel and a stride of 4, replacing the ResNet-style stem cell \cite{liu2022convnet}.

ConvNeXt also includes several advanced design elements. It uses a depthwise convolution layer with a network width of 96, maintaining the same channel configuration as the Swin Transformer, a concept called ``ResNeXt-ify." It employs an ``Inverted Bottleneck" design, similar to transformer architectures, with an expansion ratio of 4, meaning the hidden dimension of the multi-layer perceptron block is four times wider than the input dimension \cite{liu2022convnet}. Additionally, ConvNeXt uses 7x7 depth-wise convolutions within each block to capture more comprehensive spatial information. It also integrates various layer-wise micro designs, such as the Gaussian Error Linear Unit (GELU) activation function instead of ReLU, and layer normalization (LN) instead of batch normalization. These enhancements together contribute to ConvNeXt's superior performance in image recognition tasks.

In our study, we used ConvNeXt for transfer learning to identify water height in the AWD (Alternate Wetting and Drying) system. This method helps improve water management in agriculture by accurately measuring water levels. By using ConvNeXt and transfer learning, we aim to demonstrate the effectiveness and adaptability of deep learning models in practical applications.

\subsection{Convolutional Block Attention Module (CBAM)}
Attention Mechanisms is one of the modern techniques used in Neural
Networks in the field of Computer Vision. Human vision is the source of
inspiration for attention mechanisms. These mechanisms help the model focus on specific regions or features. In AWD system, it is hard to capture the feature from the above by the camera as it is influenced by various factors such as lighting conditions, water reflections and complex background structures.

To address these challenges, we incorporated different attention block modules in our pre-trained CNN model to reduce the error of the system. We experimented with Squeeze-and-Excitation block, Channel Attention Module and Spatial attention block. Convolutional Block Attention Module (CBAM) \cite{woo2018cbam} includes both Channel attention block and Spatial attention block.

\subsubsection{Channel Attention}
The channel attention mechanism in the CBAM is designed to focus on `what' is important given an input feature map. This mechanism applies GAP and global max pooling operations across the spatial dimensions of the input feature map independently to generate two different context descriptors which are $\mathbf{F}_{avg}^{c}$ and $\mathbf{F}_{max}^{c}$ (See Equation \ref{eq:channel_Favg} and Equation \ref{eq:channel_Fmax}) \cite{woo2018cbam}.

\begin{figure*} [ht]
  \includegraphics[width=\textwidth]{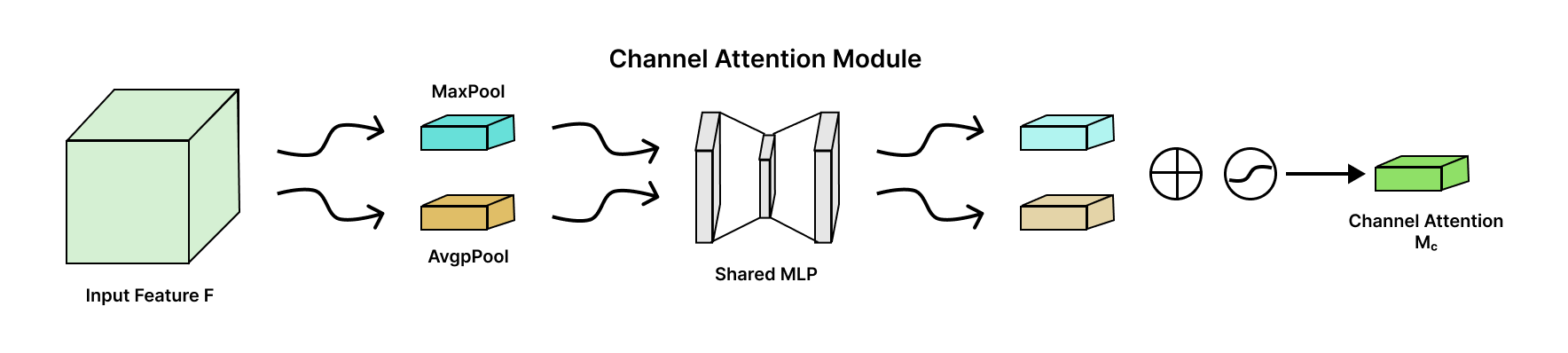}
  \caption{Channel Attention}
  \label{fig:model_arch_channel}
\end{figure*}

These descriptors are then passed through a shared network, which is a multi-layer perceptron (MLP) with one hidden layer. The outputs of the MLP for the average-pooled and max-pooled features are combined using element-wise summation, followed by a sigmoid activation function to produce the final channel attention map $\mathbf{M}_{c}$.

\begin{equation}
\mathbf{F}_{avg}^{c} = \text{MLP}(\text{AvgPool}(\mathbf{F}))
\label{eq:channel_Favg}
\end{equation}

\begin{equation}
\mathbf{F}_{max}^{c} = \text{MLP}(\text{MaxPool}(\mathbf{F}))
\label{eq:channel_Fmax}
\end{equation}

\begin{equation}
\mathbf{M}_{c}(\mathbf{F}) = \sigma(\mathbf{F}_{avg}^{c} + \mathbf{F}_{max}^{c})
\label{eq:channel_attention_eq}
\end{equation}

where $\mathbf{F}$ is the input feature map, $\sigma$ denotes the sigmoid function, $\text{AvgPool}$ and $\text{MaxPool}$ represent GAP and global max pooling operations, respectively. The resultant channel attention map $\mathbf{M}_{c}$ is then used to recalibrate the input feature map.(See Equation \ref{eq:channel_attention_eq}) \cite{woo2018cbam}. The Figure \ref{fig:model_arch_channel} shows the channel attention module.

\subsubsection{Spatial Attention}
\begin{figure*} [ht]
  \includegraphics[width=\textwidth]{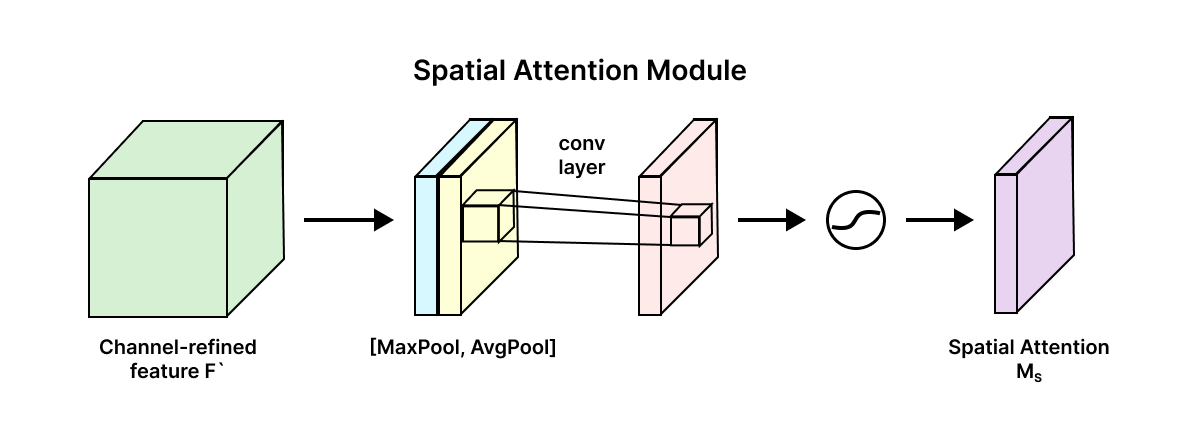}
  \caption{Spatial Attention}
  \label{fig:model_arch_spatial}
\end{figure*}

The spatial attention mechanism in CBAM focuses on `where' is important by exploiting the inter-spatial relationships of features. Unlike the channel attention mechanism, spatial attention operates on the channel-wise aggregated features. To achieve this, the input feature map is first aggregated along the channel axis using GAP and global max pooling, resulting in two 2D maps which are $\mathbf{F}{avg}^{s}$ and $\mathbf{F}{max}^{s}$ \cite{woo2018cbam}. Equation \ref{eq:spatial_Favg} and \ref{eq:spatial_Fmax} shows the calculation and Equation \ref{eq:spatial_attention_eq} shows the spatial feature maps calculation.

\begin{equation}
\mathbf{F}_{avg}^{s} = \text{AvgPool}(\mathbf{F}, \text{axis}=\text{channel})
\label{eq:spatial_Favg}
\end{equation}

\begin{equation}
\mathbf{F}_{max}^{s} = \text{MaxPool}(\mathbf{F}, \text{axis}=\text{channel})
\label{eq:spatial_Fmax}
\end{equation}

\begin{equation}
\mathbf{M}_{s}(\mathbf{F}) = \sigma(\text{Conv}([\mathbf{F}_{avg}^{s}; \mathbf{F}_{max}^{s}], \text{kernel size}=7))
\label{eq:spatial_attention_eq}
\end{equation}

where $\sigma$ denotes the sigmoid function, kernel size 7 denotes 7 $\times$ 7 kernal size and $\text{Conv}$ represents the convolution operation. The attention map $\mathbf{M}_{s}$ is used to recalibrate the input feature map spatially \cite{woo2018cbam}. The Figure \ref{fig:model_arch_spatial} shows the diagram of spatial attention module.

\section{Experimental Analysis}
In this section, we represent the experimental results and rigorous analysis of our experimented models with statistical validation.

\subsection{Dataset Description}

The dataset utilized in this study was sourced from the two location. The first location was Lalmonirhat and the second location was Gazipur in Bangladesh. The dataset includes measurements of water height at different time intervals, as well as corresponding environmental factors such as temperature, humidity, and rainfall. It was taken from a ESP-32 camera with the IoT device box which was installed in the rice field. We took both day and night pictures. The dataset was collected over a span of several months to capture the seasonal variations in water height. The height of the AWD pipe was 25 cm with the radius of 10 cm. The camera was installed in a fixed position facing the AWD pipe to capture images of the water level. We took the water height from ultrasonic sensor as ground truth for training and evaluating the water height detection model. We took images and height at the same timestamp to ensure accurate labeling of the water level in the images.

Our dataset had 1,275 data points. To make the dataset bigger and more varied, we used data augmentation. This is important for improving our model, especially because the dataset is small. We applied three types of augmentation: rotation, flipping, and brightness adjustment. These methods change how the images look, which helps the model perform better in real life. 
We randomly applied one of these augmentation methods to each image. As a result, the total number of images was increased to 2,550. This approach not only diversifies the training data but also ensures that the model can learn to recognize water height under various conditions and perspectives.

\subsection{Preprocessing}
% In this section, we detail the preprocessing steps applied to input images before subjecting them to further analysis.

% In the preprocessing pipeline, the input image is first loaded and resized to ensure uniform dimensions, followed by Gaussian blurring to reduce noise and subtle variations. The unsharp mask technique is then applied to sharpen the image, enhancing edge visibility and fine details. Conversion to grayscale simplifies processing, and Canny edge detection is utilized to identify prominent edges. Subsequently, the resulting edge map is inverted to standardize representation, followed by normalization to a floating-point range for numerical stability.

In the preprocessing pipeline, the input image is first loaded and resized to $224 \times 224$ to ensure all images have the same size. Resizing images ensures they are uniform, which is important for consistency and helps reduce the workload during training or inference. Followed by Gaussian blurring is applied to reduce noise and smooth out small variations. It smooths the image by averaging pixel values with their neighbors, which helps to suppress noise and minor artifacts that could interfere with subsequent processing steps. The unsharp mask technique is then used to sharpen the image and make edges and details more visible. It subtracts a blurred version of the image from the original image, thereby emphasizing edges and fine details, which can be critical for tasks requiring precise feature extraction. Converting the image to grayscale simplifies the processing. This reduces the image from three color channels to one, making it easier to work with and focusing on light and dark variations instead of colors. Canny edge detection is then used to find prominent edges. This method highlights important edges in the image, which helps in identifying object boundaries and structures. The edge map is then inverted to standardize how edges are represented. Inverting ensures that edges are consistently represented in all images. Finally, normalization is applied to scale the pixel values to a standard range. Normalization, often scaling values between 0 and 1, improves numerical stability and helps the machine learning algorithms work better by ensuring consistent input values.

\begin{figure*} [ht]
  \includegraphics[width=\textwidth]{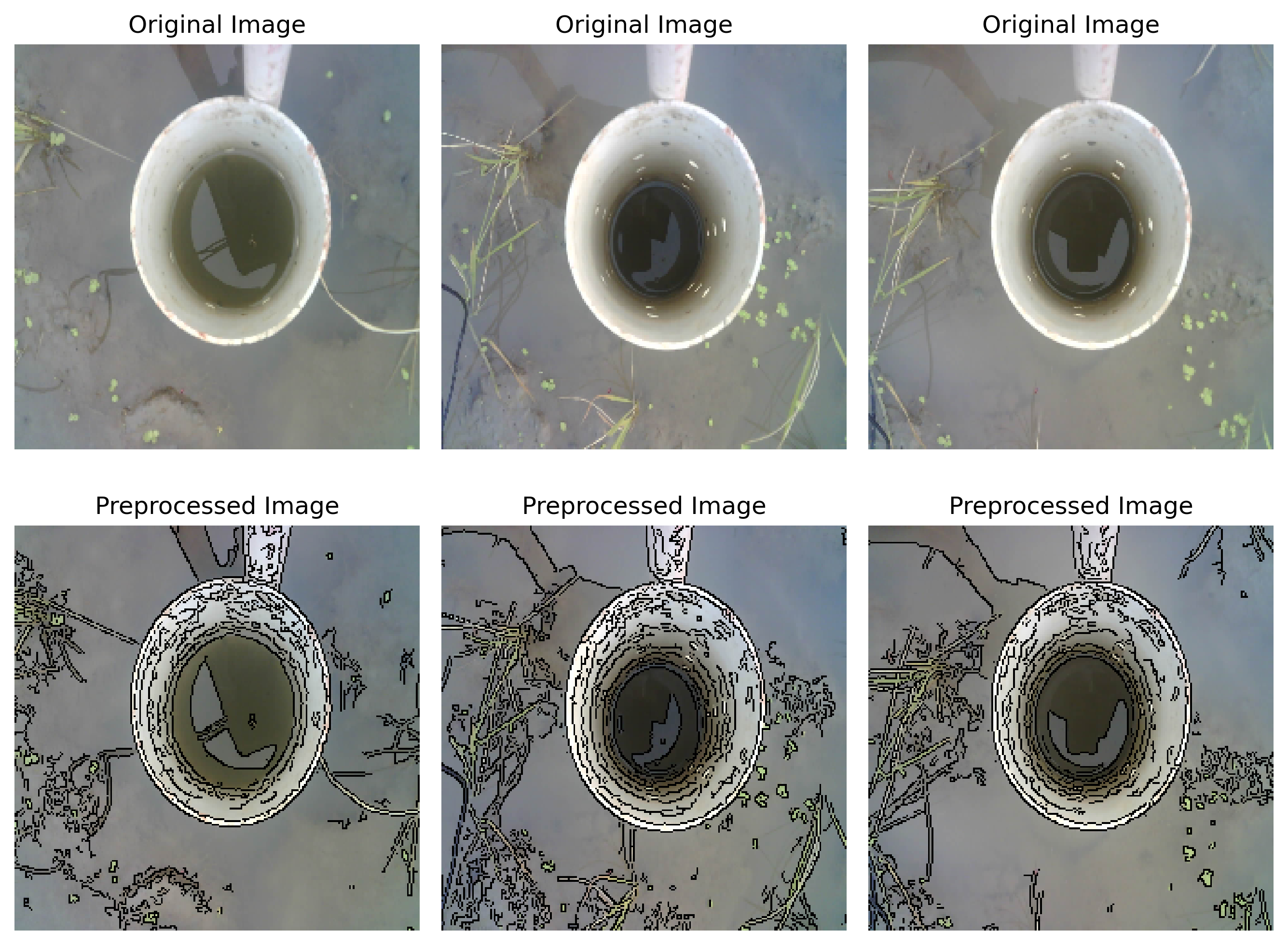}
  \caption{Original and Preprocessed Images}
  \label{fig:img_process}
\end{figure*}

\subsection{Parameters of the experiment}
To train the proposed model, we incorporated an augmented dataset, which was divided into three portions with an 80:10:10 ratio: 80\% of the images were used for training, 10\% for validation, and 10\% for testing. We initialized the ConvNeXt architecture with pre-trained weights. For our dataset, the batch size was set to 8, and the learning rate was 0.0001. We used Mean Square Error (MSE) as the loss function and the Adam optimizer to train the model over a total of 30 epochs. The experiments and image preprocessing were conducted on an HP ZBook Studio G5 laptop with 32 GB RAM, an Intel Xeon E-2176M CPU @ 2.70GHz (12 CPU cores), running the Ubuntu operating system. Additionally, we utilized the Google Colab Pro cloud service (https://colab.research.google.com), also known as `Colaboratory', to train the proposed architecture.

% We evaluated the performance of the proposed models on the water height detection task over the collected dataset. The collected dataset was divided into train, validation and test sets in the ratio of 70:15:15 respectively. 

\subsection{Evaluation Metrics}
% The performance metrics used in our experiment are Mean Squared Error(MSE) and R-squared (R2) score. RMSE determines the magnitude of error of the model and the R-squared score indicates the fitness of the model. 

In this study, we use two different metrics to quantify the experimental results: MSE and R-squared score. The expressions of the metrics are described as follows:

\begin{equation}
\text{Mean Squared Error} = \frac{1}{n} \sum_{i=1}^{n} (y_i - \tilde{y}_i)^2
\label{eq:mse}
\end{equation}

\begin{equation}
R^2 \text{ Error} = 1 - \frac{\sum_{i=1}^{n} (y_i - \tilde{y}_i)^2}{\sum_{i=1}^{n} (y_i - \bar{y})^2}
\label{eq:r2}
\end{equation}

% \subsubsection{Mean Squared Score (MSE)}
% The statistical mean squared error method calculates the average of the sum of squared distances between our target variable and the predicted values. 

% \begin{equation}
% \text{MSE} = \frac{1}{n} \sum_{i=1}^{n} (y_i - \tilde{y}_i)^2
% \label{eq:mse}
% \end{equation}

Here, \(y_i\) represents the actual value, \(\tilde{y}_i\) represents the predicted value, and \(n\) denotes the number of observations. In eq. \ref{eq:mse} the summation term \(\sum_{i=1}^{n} (y_i - \tilde{y}_i)^2\) calculates the squared differences between the actual and predicted values for each observation, and the mean of these squared differences is obtained by dividing by \(n\). This metric effectively penalizes larger errors more than smaller ones, making it a robust measure for evaluating model performance. In eq. \ref{eq:r2}, the numerator \(\sum_{i=1}^{n} (y_i - \tilde{y}_i)^2\) is the sum of the squared errors of the model, and the denominator \(\sum_{i=1}^{n} (y_i - \bar{y})^2\) is the total sum of squares (TSS), which measures the total variance in the actual data. The R\(^2\) score ranges from 0 to 1, with values closer to 1 indicating that a greater proportion of variance is explained by the model, and thus, the model has a better fit.

% \subsubsection{R-squared (R2) score}
% R-squared (R\(^2\)) score, also known as the coefficient of determination, is another key metric used to evaluate the performance of a regression model. The R\(^2\) score provides an indication of the proportion of the variance in the dependent variable that is predictable from the independent variables. It is calculated using the following equation:

% \begin{equation}
% R^2 = 1 - \frac{\sum_{i=1}^{n} (y_i - \tilde{y}_i)^2}{\sum_{i=1}^{n} (y_i - \bar{y})^2}
% \label{eq:r2}
% \end{equation}

% \(y_i\) represents the actual value, \(\tilde{y}_i\) represents the predicted value, and \(\bar{y}\) is the mean of the actual values.

\section{Result Analysis}

% \subsection{Performance Results on proposed model}

In our initial experiments, we focused on baseline CNN models without incorporating attention blocks. We selected several well-known pretrained models for our analysis, including InceptionNetV3 \cite{szegedy2016rethinking}, VGG16 \cite{simonyan2014very}, ResNet50 \cite{he2016deep}, EfficientNetB0 \cite{tan2019efficientnet}, and ConvNeXt. These models are widely recognized as benchmark architectures for image classification tasks, providing a solid foundation for evaluating performance in our study.

We tested the models with our test set. ConvNext performed better among all the pretrained models with the MSE of 0.2700 cm and R2 score of 0.9884. EfficientNetB0 showed higher MSE score which is 1.1404. The Resnet50 is another model which performs relatively well than Vgg16 and InceptionNetV3. The experimental results are shown in Table~\ref{tab:table1}.

The ConvNeXt model was tested with new field data and was decided for deployment but the model was failing to predict images accurately which had noises and cpatured in low light or comparatively blurred.

To address the shortcomings of the ConvNeXt model, particularly its difficulty in accurately predicting images with noise, low light, or blur, we incorporated attention mechanisms into the architecture. Specifically, we integrated Convolutional Block Attention Module (CBAM) and Squeeze-and-Excitation Networks (SENet) into the pretrained ConvNeXt architecture to enhance its performance on challenging image conditions. CBAM and SENet blocks have shown remarkable capabilities in enhancing feature representation after incorporating with Resnet50 and ConvNeXt architectures. These mechanisms allow the model to dynamically focus on informative regions of the image, thereby improving its ability to handle variations in the input data. Our experimental results demonstrate significant improvements when these attention layers are applied to the ConvNeXt model.

The proposed model with the integration of CBAM block, ConvNeXt model achieved an MSE of 0.2740 and R2 score of 0.9888 on the test set, outperforming the original ConvNeXT model without attention. The  results are summarized in the Table~\ref{tab:table2}.
The integration of SE-net block in the ConvNext model has also shown to improve the results with MSE of 0.2766.

The enhanced performance of the attention-based models can be attributed to their ability to adaptively focus on the most informative regions of the input images, thereby improving the overall water height prediction accuracy. Our experiments demonastrate that the incorporation of attention model such as CBAM, into a state-of-the-art ConvNeXt architecture can significantly improve the  water height estimation capabilities in the AWD system
% \usepackage{multirow}

\begin{table}[]
\centering
\caption{Pretrained Models Result}
\vspace{0.5em} % Add vertical space for line break
\begin{tabular}{|c|ll|}
\hline
\multirow{2}{*}{Pretrained Models} & \multicolumn{2}{c|}{Test Results}      \\ \cline{2-3} 
                                   & \multicolumn{1}{c|}{MSE}    & R2 Score \\ \hline
InceptionNetV3                     & \multicolumn{1}{c|}{0.4807} & 0.9798   \\ \hline
Vgg16                              & \multicolumn{1}{c|}{0.8955} & 0.9624   \\ \hline
Resnet50                           & \multicolumn{1}{c|}{0.4609} & 0.9806   \\ \hline
EfficientNetB0                     & \multicolumn{1}{c|}{1.1404} & 0.9522   \\ \hline
ConvNeXt                           & \multicolumn{1}{c|}{0.2778} & 0.9863   \\ \hline
\end{tabular}
\label{tab:table1}
\end{table}

% \usepackage{multirow}
\begin{table}[]
\centering
\caption{Attention-based Models Result}
\vspace{0.5em}
\begin{tabular}{|c|ll|}
\hline
\multirow{2}{*}{Attention based Models} & \multicolumn{2}{c|}{Test Results}      \\ \cline{2-3} 
                                        & \multicolumn{1}{c|}{MSE}    & R2 Score \\ \hline
Resnet50+CBAM                           & \multicolumn{1}{c|}{0.7652} & 0.9679   \\ \hline
SeResnet50                              & \multicolumn{1}{c|}{0.3689} & 0.9845   \\ \hline
ConvNeXt+CBAM                           & \multicolumn{1}{c|}{0.2740} & 0.9885   \\ \hline
SeConvNeXt                              & \multicolumn{1}{c|}{0.2766} & 0.9884   \\ \hline
\end{tabular}
\label{tab:table2}
\end{table}

\subsection{Ablation study}
To determine the effect of image preprocessing, attention mechanism and source of data, we trained and fine tuned our model. We experimented in various set of combinations  to select the best available configurations for the proposed model.

\begin{table}[]
\centering
\caption{Ablation study for selecting the architectures}
\vspace{0.5em}
\resizebox{\textwidth}{!}{
\begin{tabular}{|l|l|c|c|c|}
\hline
\multicolumn{1}{|c|}{\textbf{Data Source}} & \multicolumn{1}{c|}{\textbf{Model Name}} & \textbf{Image Preprocessing} & \textbf{MSE}    & \textbf{R2 Score} \\ \hline
\multirow{4}{*}{Lab Data}                  & \multirow{2}{*}{ConvNeXt}              & Yes                          & 0.021           & 0.9910            \\ \cline{3-5} 
                                           &                                          & No                           & 0.018           & 0.9922            \\ \cline{2-5} 
                                           & \multirow{2}{*}{Proposed Model}          & Yes                          & 0.022           & 0.9910            \\ \cline{3-5} 
                                           &                                          & No                           & \textbf{0.015}  & \textbf{0.9980}   \\ \hline
\multirow{4}{*}{Field Data}                & \multirow{2}{*}{ConvNeXt}              & Yes                          & 0.2770          & 0.9863            \\ \cline{3-5} 
                                           &                                          & No                           & 0.2880          & 0.9872            \\ \cline{2-5} 
                                           & \multirow{2}{*}{Proposed Model}          & Yes                          & \textbf{0.2740} & \textbf{0.9885}   \\ \cline{3-5} 
                                           &                                          & No                           & 0.2890          & 0.9875            \\ \hline
\end{tabular}
}
\label{tab:ablation}
\end{table}

We experimented with different baseline pretrained CNN architectures and found the best results in ConvNeXt architecture. We considered this architecture as our base model. In the ablation study, we compared the performance of the base model and the proposed model on both Lab Data and Field Data, evaluating the impact of image preprocessing on MSE and R2 scores. For Lab Data, the proposed model without image preprocessing achieves the lowest MSE of 0.015 and the highest R2 score of 0.9980, indicating superior performance over the base model which has the MSE of 0.018 and R2 score of 0.9922. In contrast, for Field Data, the proposed model with image preprocessing performs marginally better than without preprocessing, achieving an MSE of 0.2740 and an R² of 0.9885. In Table~\ref{tab:ablation}, the result shows that the proposed model is highly effective even without preprocessing for Lab Data and after applying preprocessing to Field Data, it can provide a marginal performance enhancement.
\subsection{Discussion}

\begin{figure}[ht]
  \centering	
  \includegraphics[width=\textwidth]{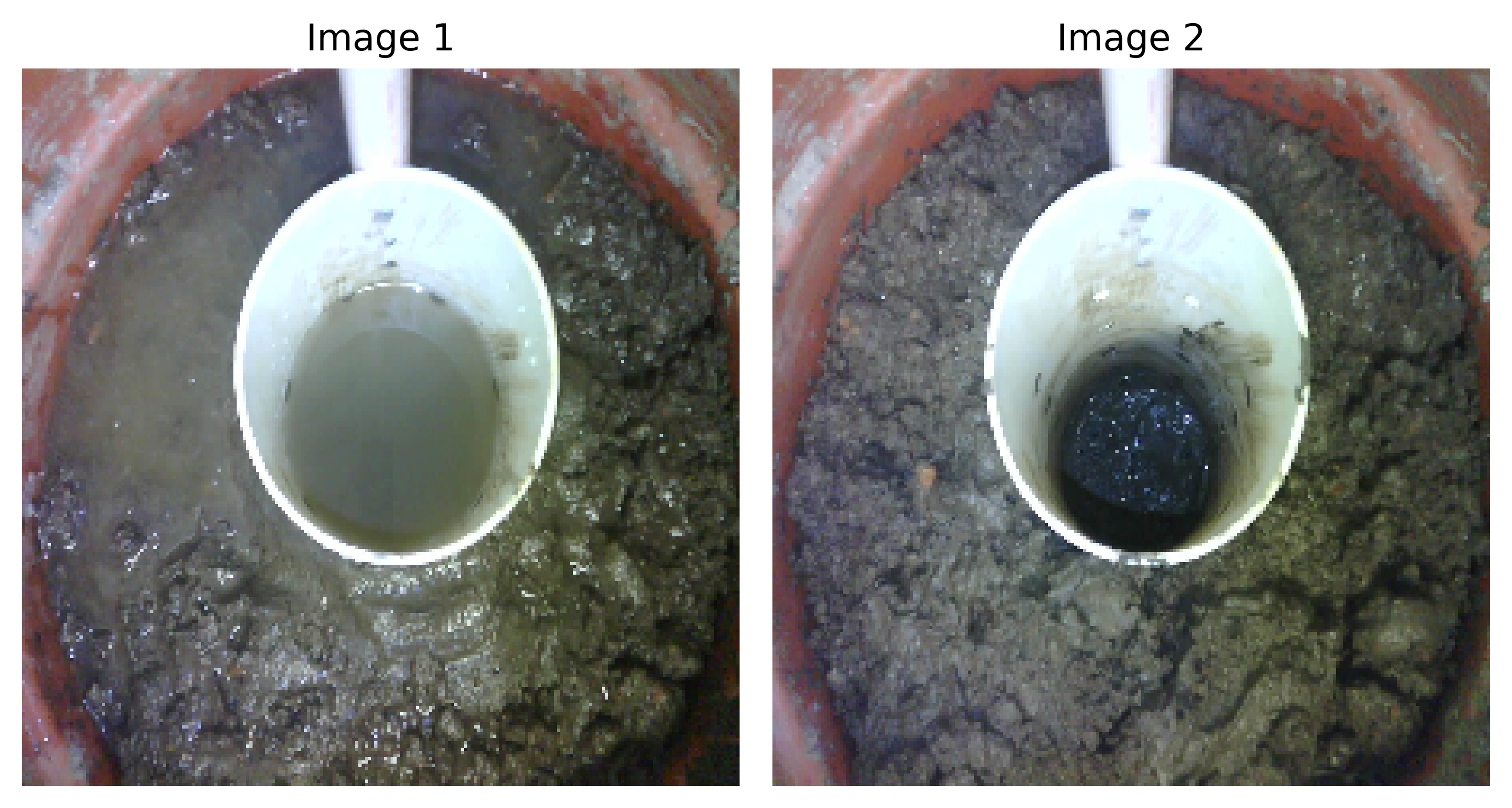}
  \caption{Data Taken in the Laboratory.}
  \label{fig:lab_setup}
\end{figure}

The results indicate that the proposed attention-based models outperformed the baseline CNN models in the water height estimation task. Prior to implementing attention-based solutions, we tested our hypothesis by collecting indoor data using the experimental setup illustrated in Figure~\ref{fig:lab_setup}.

We initially tested our model using indoor data collected in a controlled environment, where factors such as lighting, background conditions, air flow, and temperature were regulated. The indoor test data comprised images with clear water surfaces, free from noise or disturbances. In this controlled setting, the baseline CNN model without attention blocks performed well. However, when applied to outdoor data, the model's performance degraded significantly due to environmental factors and noise. The field setup presented several challenges, including thunderstorms, strong air flow, extreme sunlight fluctuations, water ripples, and reflections, all of which negatively impacted the model's accuracy on field data.

To better understand the effects of airflow and temperature, we tested our lab setup outdoors, focusing on the fluctuations of ultrasonic sensors in field conditions. Under normal airflow and temperature, the ultrasonic sensor readings fluctuated between 0.5 and 1 cm. However, during rain, thunderstorms, or other extreme environmental conditions, the sensors failed to provide accurate measurements.

To address these issues, we conducted field experiments at the Bangladesh Rice Research Institute (BRRI), where we manually measured water levels using a scale and took photos of the water surfaces from a fixed camera location. We cross-verified the ultrasonic sensor readings with manual measurements to ensure reliability. After confirming consistent results, we proceeded to collect images under various environmental conditions, such as high sunlight, water reflections, shadows, and rain.

We expanded our dataset by collecting data from three different locations—Lalmonirhat, Bogura, and Gazipur—creating a diverse dataset that includes various geographical and environmental factors. This dataset enhances the reliability and robustness of our solution, making it better suited for real-world applications.

\section{Conclusion}
In this paper, we introduced a robust water height estimation system using a CNN enhanced with attention mechanisms. We also proposed a novel dataset specifically designed for AWD systems, offering a reliable solution for water conservation in irrigation.

Our experimental results show that incorporating attention mechanisms like CBAM and SE-net into the ConvNeXT architecture significantly improves water height estimation, even under challenging environmental conditions. The proposed system, which automates water height measurement using camera technology, offers high durability, reliability, and lower maintenance costs compared to traditional ultrasonic sensors. This solution can be deployed in the field to accurately measure water height and automatically adjust irrigation systems, contributing to more efficient water management in agriculture. It also supports deployment on both edge devices and cloud platforms.

While the results are promising, this study has some limitations. The model’s performance in extreme weather conditions, which were not extensively tested, could vary. Additionally, although the dataset is comprehensive, it may not cover every environmental scenario in diverse agricultural settings. In future work, we plan to expand the dataset to include a wider variety of conditions and explore additional attention mechanisms to further improve estimation accuracy.

\section*{Acknowledgement}
The research work was funded and supported by Institute for Advanced Research (IAR), an institute within United International University (UIU) and Nodes Digital Limited (Project Code: UIU-IAR-01-2022-SE-09).

\bibliographystyle{elsarticle-num} 
\bibliography{cas-refs}

\end{document}